\documentclass[10pt, conference, letterpaper]{IEEEtran}
\usepackage[ruled,linesnumbered,vlined]{algorithm2e}
\usepackage{verbatim}
\usepackage{amsfonts}
\usepackage{epsfig}
\usepackage{algorithmic}
\usepackage{multicol}
\usepackage{multirow}
\usepackage{amsmath}
\usepackage{amssymb}
\usepackage{subfigure}
\usepackage{subfig}
\usepackage{textcomp}
\usepackage{tikz}
\usepackage{pgf}
\usepackage{float}
\usepackage{graphicx}
\usepackage{comment}
\usepackage{mathtools}

\usepackage{url}      

\usepackage{diagbox}
\usepackage{array}
\newcolumntype{M}[1]{>{\centering\arraybackslash}m{#1}}

\makeatletter
\def\url@leostyle{%
  \@ifundefined{selectfont}{\def\UrlFont{\sf}}{\def\UrlFont{\small\bf\ttfamily}}}
\makeatother
\begin{document}

\title{Data Dropout: Optimizing Training Data for Convolutional Neural Networks}

\author{\IEEEauthorblockN{Tianyang Wang} 
\IEEEauthorblockA{Austin Peay State University\\
toseattle@siu.edu}

\and
\IEEEauthorblockN{Jun Huan} 
\IEEEauthorblockA{Baidu Research\\
huanjun@baidu.com}

\and
\IEEEauthorblockN{Bo Li} 
\IEEEauthorblockA{University of Southern Mississippi\\
bo.li@usm.edu}

}

\providecommand{\keywords}[1]{\textbf{\textit{Keywords---}} #1}

\maketitle

\begin{abstract}

Deep learning models learn to fit training data while they are highly expected to generalize well to testing data. Most works aim at finding such models by creatively designing architectures and fine-tuning parameters. To adapt to particular tasks, hand-crafted information such as image prior has also been incorporated into end-to-end learning. However, very little progress has been made on investigating how an individual training sample will influence the generalization ability of a model. In other words, to achieve high generalization accuracy, do we really need all the samples in a training dataset? In this paper, we demonstrate that deep learning models such as convolutional neural networks may not favor all training samples, and generalization accuracy can be further improved by dropping those \emph{unfavorable} samples. Specifically, the influence of removing a training sample is quantifiable, and we propose a \emph{Two-Round Training} approach, aiming to achieve higher generalization accuracy. We locate \emph{unfavorable} samples after the first round of training, and then retrain the model from scratch with the reduced training dataset in the second round. Since our approach is essentially different from fine-tuning or further training, the computational cost should not be a concern. Our extensive experimental results indicate that, with identical settings, the proposed approach can boost performance of the well-known networks on both high-level computer vision problems such as image classification, and low-level vision problems such as image denoising.

\end{abstract}


\section{Introduction}
\label{introduction}

Pioneered by AlexNet \cite{krizhevsky2012imagenet}, deep learning models such as convolutional neural networks (CNNs) gain remarkable success in solving computer vision problems \cite{simonyan2014very, szegedy2015going, long2015fully, he2016deep, huang2017densely, he2017mask}. One major research interest is to design powerful architectures to extract more distinguished features from data. Such examples include ResNet \cite{he2016deep} and DenseNet \cite{huang2017densely}. Moreover, to ease the training of deep networks and to alleviate over-fitting, a lot of techniques were proposed. Typical works include batch normalization (BN) \cite{ioffe2015batch}, and dropout \cite{krizhevsky2012imagenet}. To enhance the expressive power of CNNs, advanced non-linear activation functions were well-studied, such as ReLU \cite{nair2010rectified}, ELU \cite{clevert2015fast}, and SELU \cite{klambauer2017self}. Besides, a lot of domain specific techniques were also developed to further fine-tune networks on specific applications. Detailed discussion of such techniques is beyond the scope of this work. 

Another big bottleneck in deep learning is lack of real data. Deep CNN training must rely on a high volume of data. Unfortunately, this condition is not always satisfied in real scenarios. Therefore, data augmentation becomes a feasible and indispensable approach to increase data diversity. Typical ways such as image rotation, flipping, and shifting are widely used in data pre-processing. Recently, generative adversarial network (GAN) \cite{goodfellow2014generative} are also widely utilized to generate synthetic data which cannot even be differentiated by discriminative models or human beings.

However, though a lot of efforts have been made on the aforementioned aspects, it is still unclear how an individual training sample will influence the generalization accuracy of a network. To clarify this problem, let us consider two questions. (1) Given a network and its training dataset, can we drop several training samples so that generalization accuracy can be improved? (2) If so, how can we leverage the model to fit a subset of the given training dataset? Our work demonstrates that there exist such training samples which we call \emph{unfavorable} training samples. We propose a \emph{two-round training} approach to improve CNN's generalization accuracy by dropping those samples, and we name the dropping step as \emph{data dropout}, which is a scheme for training data optimization. Specifically, we train a network with a given training set in the first round, then for each training sample, we compute the influence of removing it on the loss across all validation samples. If the influence value is positive, implying that its removal will reduce the whole validation error, we will drop that training sample. Therefore, the training set can be rebuilt. In the second round, we use the reconstructed training set to retrain the network from scratch to obtain a new trained model which will be used for testing. To make our approach more general, we measure the influence of each training sample based on a validation set instead of a testing set, because testing data is usually unavailable during training stage. If no validation set is given originally, one can randomly separate a group of samples from the given training set as validation samples. Even though a network will see fewer training samples due to the removing of \emph{unfavorable} samples, extensive experiments still demonstrate that our \emph{data dropout} scheme implemented by the \emph{two-round training} approach can further boost performance of the state-of-the-art networks, such as ResNet \cite{he2016deep} and DenseNet \cite{huang2017densely}. Despite of simpleness, our approach does not rely on particular networks and training configurations. The only prerequisite is a network model that can fit original training data for specific tasks. Therefore, it is convenient to apply our approach with existing CNN models.

It is worth noting that our approach is essentially different from further training or fine-tuning, because we retrain the model from scratch in the second round. The model trained in the first round is only utilized to compute influence values for training data optimization. 


The main contributions of this work can be generalized into three folds.
\begin{itemize}
\item Firstly, we propose \emph{data dropout} scheme to optimize training set by removing \emph{unfavorable} samples.
\item Secondly, we design a \emph{two-round training} approach to leverage \emph{data dropout} to improve generalization accuracy.
\item Thirdly, we conduct extensive experiments to demonstrate the effectiveness and generality of our approach in boosting the performance of existing CNN models that were designed for diverse computer vision problems such as image classification and image denoising.
\end{itemize}

\section{Related Work}
\label{relatedwork}

In this section, we review related literature and several benchmark datasets that will be used in our experiments. We also briefly introduce image denoising, a low-level computer vision problem that will be adopted as an example application in the experiments section.

\subsection{Most Related Research}

Our work is partially inspired by \cite{koh2017understanding}, but it is worth noting that our work is different in the following aspects. Firstly, the authors' work mainly concentrated on model behavior while our work focuses on optimizing training data such that we can achieve even better performance with existing network models. Secondly, they studied the feasibility of approximating the influence of removing a training sample on the loss at a testing sample, however, they did not establish a criteria for \emph{unfavorable} samples, while we explicitly propose this criteria in Section \ref{criteria}. 

\subsection{Image Classification}
Image classification has been a classical task to evaluate CNNs. The well-known models such as All-CNN \cite{springenberg2014striving}, ResNet \cite{he2016deep}, and DenseNet \cite{huang2017densely} were originally proposed for this task. In our experiments, we adopt four widely used datasets consisting of the two CIFAR datasets \cite{krizhevsky2009learning}, the SVHN (Street View House Numbers) dataset \cite{netzer2011reading}, and the ImageNet dataset \cite{deng2009imagenet}. The CIFAR-10 and the  CIFAR-100 datasets contain 10 and 100 classes of color images, respectively. There are 50,000 training samples and 10000 testing samples, all in size $32\times32$. The SVHN dataset contains 73,257 training images and 26,032 testing images belonging to 10 classes. There are also 531,131 images in the additional training set. All the images have a dimension of $32\times32$. The ImageNet dataset contains 1.28 million images for training, 50000 images for validation, and 100000 images for testing. There are 1000 classes in total. In real practice, all these color images can be cropped to a fixed size, such as $224 \times 224$.


\subsection{Image Denoising}
\label{imgdenoising}
Image denoising has been a long term open and challenging low-level computer vision problem. The degradation is usually modeled as $y$ = $x$ + $\epsilon$, where $x$ denotes latent clean image, $\epsilon$ additive Gaussian noise and $y$ corrupted observation. In addition to image prior method, discriminative learning based approaches have been widely applied on denoising research. Typical works include MLP \cite{burger2012image}, CSF\cite{schmidt2014shrinkage}, NLNet \cite{lefkimmiatis2016non}, and DnCNN \cite{zhang2017beyond}, which presented very competitive results. Unlike other methods that aim to learn latent clean image $x$ directly, DnCNN leveraged residual learning to learn noise $\epsilon$. Clean image $x$ can be restored by subtracting the learned noise $\epsilon$ from corrupted observation $y$. For reasonable comparison, we will directly adopt DnCNN model and its initial configurations in our experiments.

\section{The Proposed Method}
\label{Method}

To ease the discussion, we start by defining several notations. Let $x_{i}$ ($i=1,...,n$) denote a training sample. Let $f_{\theta}(x)$ denote a model such as CNN with input $x$, $L(f_{\theta}(x))$ the loss, and $I_{loss}(x, x_{j})$ the influence of removing a training sample $x$ on the loss at a validation sample $x_{j}$. Here, $j=1,...,k$ and $k$ equals the number of validation samples. The goal of training is to learn a set of parameters $\theta = \arg\min_{\theta} \frac{1}{n} \sum_{i=1}^n L(f_{\theta}(x_{i}))$, where $L$ can be typical loss functions, such as softmax loss, mean squared error (MSE), and $L_2$ loss, etc.


\subsection{Influence Computation}
\label{computeInf}
According to the approximation theory discussed in \cite{koh2017understanding}, $I_{loss}(x, x_{j})$ can be defined as below,
\begin{equation}
  I_{loss}(x, x_{j})=-\nabla_\theta L(f_{\theta}(x_{j}))^\top H_{\theta}^{-1} \nabla_\theta L(f_{\theta}(x)),
\end{equation}
where $H_{\theta}^{-1}=\frac{1}{n} \sum_{i=1}^n \nabla^2_\theta L(f_{\theta}(x_{i}))$ is the Hessian 
and assumed to be positive definite. In our work, for each training sample $x$, we compute its influence on loss value over all validation samples instead of testing samples since testing data should be invisible until testing phase. Hence, the total influence is $\sum_{j}I_{loss}(x, x_{j})$. 
To compute $I_{loss}(x, x_{j})$, which can be rewritten as follows,

\begin{equation}
  I_{loss}(x, x_{j})=-s_{j}\cdot \nabla_\theta L(f_{\theta}(x)),
\end{equation}
where $s_{j}=\nabla_\theta L(f_{\theta}(x_{j}))^\top H_{\theta}^{-1}$, we approximate $s_{j}$ by using implicit Hessian-vector products. Stochastic estimation technique can be used to solve such an approximation problem. One can refer to \cite{koh2017understanding} for more details. In our experiments, we note that for each training sample $x$, computing $\sum_{j}I_{loss}(x, x_{j})$ over all $x_{j}s$ at one time is still computing-intensive, hence we slightly change the order of computation which can greatly improve the efficiency. We will detail the implementation tips in Section \ref{impldetail}.

\subsection{Data Dropout Criteria}
\label{criteria}
Once $I_{loss}(x, x_{j})$ can be computed, we will be able to compute the total influence $\sum_{j}I_{loss}(x, x_{j})$ across all validation samples, which is used to approximate the following,
\begin{equation}
  \sum_{j} L(f_{\theta}(x_{j}))-L(f_{\theta'}(x_{j})), 
\end{equation}
where ${\theta'}$ is defined as $\theta' = \arg\min_{\theta} \frac{1}{n} \sum_{x_{i}\neq{x}} L(f_{\theta}(x_{i}))$.
In practice, we expect $\sum_{j} L(f_{\theta}(x_{j}))-L(f_{\theta'}(x_{j}))>0$, which implies that removing a training sample $x$ can decrease total validation loss, hence it is equivalent to have $\sum_{j}I_{loss}(x, x_{j})>0$. Therefore, we set the criteria of \emph{data dropout} as follows:
$\forall{x}$, if $\sum_{j}I_{loss}(x, x_{j})>0$, $x$ will be dropped from training set, otherwise, it will be kept. The dropped $x$ is named as \emph{unfavorable} sample in this context.

It is important to note that we utilize validation loss to closely reflect potential testing loss when performing data dropout. This makes sense because training data is commonly assumed to have similar data distribution as potential testing data in machine learning. Otherwise, the problem may lie in the category of transfer learning \cite{pan2010survey} that is beyond the scope of this work. In this context, our validation data is usually separated from original training data, and it can be assumed having similar distribution as potential testing data. Since $\sum_{j}I_{loss}(x, x_{j})>0$, we will have $\sum_{k}\sum_{j}I_{loss}(x_{k}, x_{j})>0$, where $x_{k}$ denotes an \emph{unfavorable} training sample. Thus we will obtain 
\begin{equation}
  \sum_{k}\sum_{j} L(f_{\theta}(x_{j}))-L(f_{\theta'}(x_{j}))>0, 
\end{equation}
and it indicates that testing loss can be reduced by removing \emph{unfavorable} training samples. In addition, all \emph{unfavorable} samples will be dropped at one time and thus the network parameters will be updated at one time. Therefore, the removal of each individual \emph{unfavorable} sample is independent to each other. 

\subsection{Two-Round Training}
\label{tr-train}
As analyzed above, for an individual training sample $x$, we can compute $\sum_{j}I_{loss}(x, x_{j})$, where $x_j$ is a validation sample. As a result, we want to examine each training sample to decide whether to drop or keep it. In conventional learning, once training is done, the learned parameters are fixed, hence testing error rates cannot be changed. Therefore, to make use of the computed influence to further decrease testing error rates, we propose a \emph{two-round training} approach.

In the first round, we choose an arbitrary network which is suitable for a given task, and setup training configurations according to conventional practices, such as ResNet \cite{he2016deep} for image classification. We train the model, and obtain $f_{\theta}(\cdot)$ when training is done. Here, $\theta$ denotes the learned network parameters. Then, for each training sample $x_{i}$, we compute $\sum_{j}I_{loss}(x_{i}, x_{j})$, the influence of removing $x_{i}$ on the loss over all validation samples, and remove \emph{unfavorable} $x_{i}s$ according to the criteria of \emph{data dropout}. Thus, a new training set can be rebuilt. In the second round, we use the same network and the initial configurations as the first round, but feed the reconstructed training set to the model, and retrain it. When this round of training is complete, the resulting model $f_{\theta'}(\cdot)$ is adopted as the final model for testing. Since the network is trained on the optimized training set in the second round, the learned parameters $\theta'$ are quite different from $\theta$, which are learned from the first round. We generalize our approach in the \textbf{Algorithm}.

\begin{algorithm}[tp]
\SetKwData{P_u}{P_u}\SetKwData{P_v}{P_v}\SetKwData{P_i}{P_i}\SetKwData{P_j}{P_j}
\SetKwData{T1}{t1}\SetKwData{T2}{t2} \SetKwData{E}{E}
\SetKwData{Left}{left}\SetKwData{This}{this}\SetKwData{Up}{up}
\SetKwInOut{Input}{Input}\SetKwInOut{Output}{Output}
\Input{\\
${f}$: network, $\theta_{0}$: initial parameters\; 
$\mathcal{X}$: training set, $\mathcal{V}$: validation set\; 
$i$/$j$: training/validation sample index\;
$len$($\cdot$): total number of samples in a dataset ``$\cdot$"\;}
\Output{\\
${\theta}$: parameters trained based on $\mathcal{X}$\;
$\mathcal{X'}$: optimized training dataset\;
$\theta'$: optimized parameters trained based on $\mathcal{X'}$\;
}
\Begin{
Train $f_{\theta_{0}}$ on $\mathcal{X}$ to obtain $f_{\theta}$\;
\For{$i\leftarrow 1$ \KwTo $len$($\mathcal{X}$)}{
\For{$j\leftarrow 1$ \KwTo $len$($\mathcal{V}$)}{ use $f_{\theta}$ to
\textbf{compute} $I_{loss}(\mathcal{X}(i)$, $\mathcal{V}(j))$;
}
\If{$\sum_{j}I_{loss}(\mathcal{X}(i)$, $\mathcal{V}(j))>0$}
   {\textbf{remove} $\mathcal{X}(i)$ from $\mathcal{X}$;}
}   
$\mathcal{X'}$ is obtained\;
Train $f_{\theta_{0}}$ on $\mathcal{X'}$ to obtain $f_{\theta'}$\;
}
\caption{Two-round training approach}
\label{algo_iterative}
\end{algorithm} 

\begin{figure}
    \centering
    \includegraphics[width=0.4\textwidth]{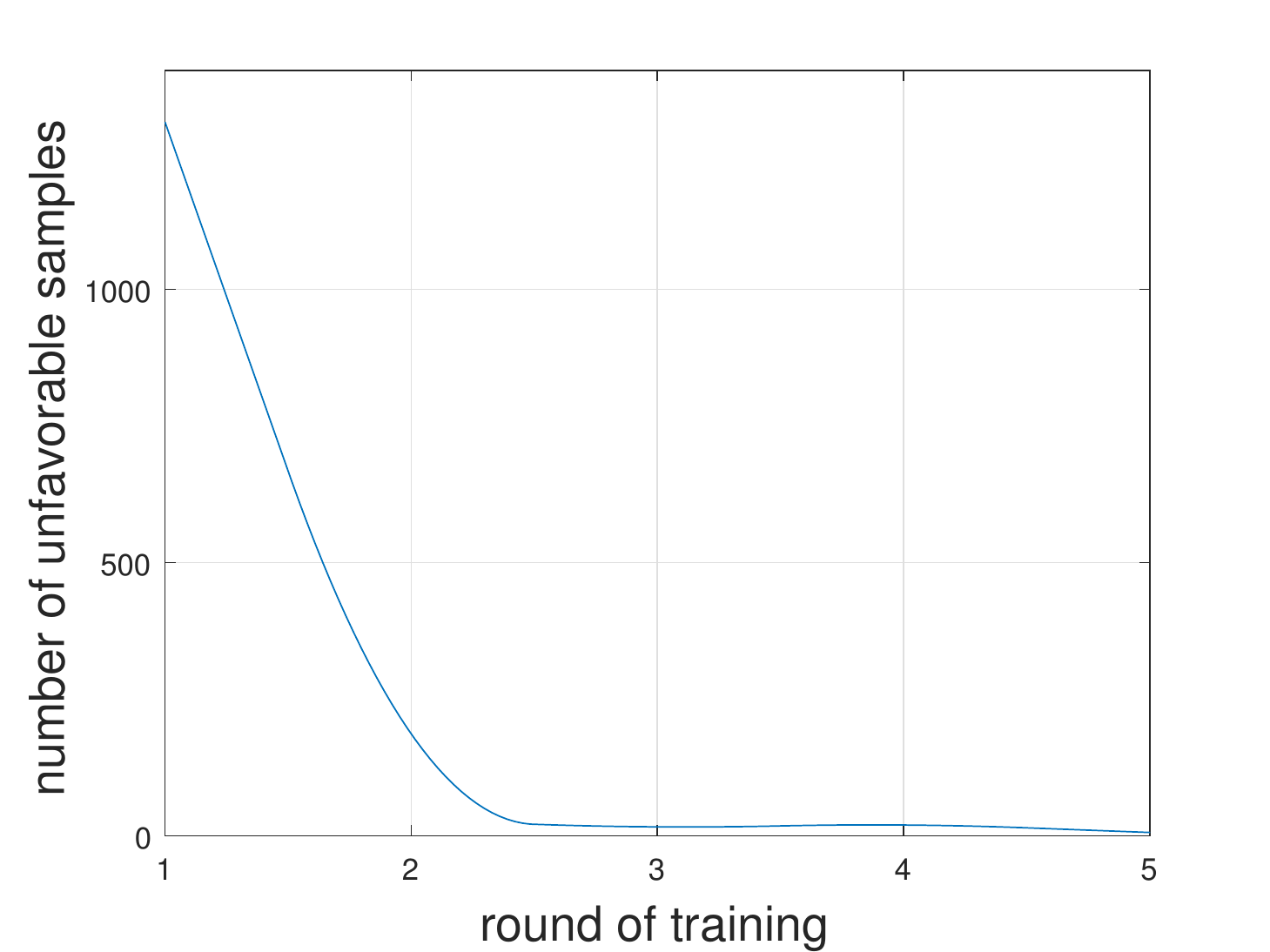}
    \caption{The number of unfavorable samples found after each round of training. It is based on training the ResNet-20 \cite{he2016deep} on the CIFAR-10 dataset without data augmentation.}
    \label{roundsample}
\end{figure}

Here, we briefly discuss the appropriate number of training rounds. In general, after the second round of training, we can still find a few \emph{unfavorable} samples. For instance, in the CIFAR-10 classification with the ResNet-20 \cite{he2016deep}, we show the amount of located \emph{unfavorable} samples after each round of training in Figure \ref{roundsample}. 
 
It can be seen that the curve is nearly monotonic, and there are much fewer \emph{unfavorable} training samples left after the first round of training. This fact indicates that it is not necessary to perform more rounds of training in order to locate more \emph{unfavorable} samples. On the other hand, more rounds will be computing-intensive, which is undesirable in deep learning. In fact, we empirically observe that two rounds of training is sufficient to improve generalization accuracy. Therefore, our approach is \emph{two-round} based considering both accuracy and efficiency. The first round is to train a model which is used for locating \emph{unfavorable} samples and the second round is to train the same network from scratch on the optimized training set for testing purpose.

\subsection{Implementation Tips}
\label{impldetail}
According to the \textbf{Algorithm} and the analysis in Section \ref{tr-train}, for each training sample $x_{i}$, $\nabla_\theta L(f_{\theta}(x_{i}))$ will be fixed and $s_{j}=\nabla_\theta L(f_{\theta}(x_{j}))^\top H_{\theta}^{-1}$ needs to be computed multiple times across all validation samples. However, approximating $s_{j}$ is more computing-intensive than computing $\nabla_\theta L(f_{\theta}(x_{i}))$. Therefore, in implementation of the \textbf{Algorithm}, we firstly fix $s_{j}$ and compute $\nabla_\theta L(f_{\theta}(x_{i}))$ for all $x_{i}s$, and then repeat this for all $x_{j}s$. In this way, we have only $k$ (the total number of $x_{j}s$) times of approximating $s_{j}$. Otherwise, for all $x_{i}s$, computing $\sum_{j}I_{loss}(x_{i}, x_{j})$ first will need $n \times k$ times of $s_{j}$ approximations, where $n$ is the number of $x_{i}s$. With this optimization, there will be $k$ values for each training sample $x_{i}$ in the end. Then, we sum these values to obtain the influence of removing each training sample on the loss at all validation samples. Although this optimization does not change the number of iterations in the \textbf{Algorithm}, it can greatly reduce the number of approximation operations.

\subsection{Difference from \textquoteleft Leave-one-out\textquoteright ~retraining}
It is worth noting that our approach is essentially different from \textquoteleft leave-one-out\textquoteright ~(LOO) retraining. For each training sample $x_{i}$, to compute the influence on validation loss, LOO needs to retrain the network by removing $x_{i}$ from the training set. Hence it needs $n$ times of retraining to investigate all training samples, which is not feasible in deep learning. Our approach, instead, computes the influence on validation loss for all $x_{i}s$ at one time after the first round of training.

Our training method looks like a fine-tuning technique. However, it is essentially different from traditional fine-tuning techniques. The reason is that in the second round of training we restore the network to its initial configurations. In fact, result of the first round of training will be completely discarded when we start the second round of training since that result is only necessary for \emph{data dropout}, which optimizes training data.

\subsection{Generality}
In spite of simpleness, our approach does not rely on particular models or applications. The model to be used to solve a domain-specific problem can be either an existing model or a customized network. There is no restriction on hyper-parameter settings, either. One only needs to follow a \emph{train-drop-train} manner to achieve further improvement on testing accuracy of their selected networks.
In addition, our approach is still applicable when there is no validation data, because we can choose some training samples or use another dataset for validation purpose. But the selected validation data should have similar distribution as potential testing data. This scheme has been proved to be effective in our experiments. 

Moreover, our approach can not only improve the state-of-the-art baselines as shown in the experiments, but also improve the performance of an arbitrary model even the model is simpler in structure. For instance, the All-CNN model \cite{springenberg2014striving} has simpler structure than the DenseNet \cite{huang2017densely} and it cannot give the state-of-the-art baselines for image classification problem, but its performance can be still boosted by employing the proposed scheme as shown in Table \ref{classification}.

\section{Experiments}
\label{experiment}

To validate the effectiveness of \emph{data dropout} and \emph{two-round training} approach, we conduct extensive experiments which include image classification, and image denoising. We choose well-known networks for each task, and follow common practices to train the networks for reasonable evaluations. All experiments are implemented in TensorFlow \cite{tensorflow2015-whitepaper} with Keras API \cite{chollet2015keras}.

In all our experiments, after removing \emph{unfavorable} training samples, we do not add additional samples to the training sets and keep the initial batch size unchanged, thus allowing \textquoteleft not-full batch\textquoteright. Moreover, to better estimate the influence of each training sample, data augmentation is turned off in the first round of training and only used in the second round for all the experiments except the SVHN, in which we turn off data augmentation for both rounds to follow common practices.

\subsection{Image Classification}
For the CIFAR-10 and the CIFAR-100 datasets, we separate 5,000 images as validation data from the training set, and the remaining 45,000 images are used for training. In the second round of training, horizontal flippings and translations are adopted for data augmentation.

For the SVHN dataset, we constitute the validation set with 4,000 images from the training set and 2,000 images from the additional training set. These images are evenly sampled from 10 classes. We pre-process the images by subtracting the mean and dividing the standard deviation.

By adopting our approach, we re-evaluate the following three well-known networks, which focus on image classification problem: ResNet \cite{he2016deep}, DenseNet \cite{huang2017densely}, and All-CNN \cite{springenberg2014striving}. We directly use the models without changing the architectures. One can refer to the original papers for architecture details. For each round of training, we adopt the \emph{MSRA} method \cite{he2015delving} to initialize parameters for ResNet and DenseNet, and the \emph{Xavier} method \cite{glorot2010understanding} to initialize All-CNN.

\vspace{6pt}
\noindent\textbf{ResNet.} To have reasonable comparison, we follow the practices in \cite{huang2016deep} for ResNet evaluation. Firstly, we re-evaluate ResNet-110 with our approach on the two CIFAR datasets. The model is trained with stochastic gradient descent (SGD) optimizer with a mini-batch size of 128, weight decay of 0.0001 and momentum of 0.9. The initial learning rate is set to 0.1, and reduced to 0.01 and 0.001 at epoch 250 and 375 out of 500 epochs, respectively. Secondly, we re-evaluate ResNet-152 on the SVHN dataset. We train the model for 50 epochs, and the learning rate is reduced to 0.01 and 0.001 at epoch 30 and 35 respectively, from the initial value of 0.1. Other hyper-parameter settings keep unchanged as in the CIFAR experiments.

\vspace{6pt}
\noindent\textbf{DenseNet.} Although DenseNet has several versions, we choose to re-evaluate the basic version (DenseNet-40) which has no bottleneck layers or compression. There are 16 filters in the initial layer and the growth rate is set to 12. We train the model for 300 and 40 epochs on the CIFAR and the SVHN dataset, respectively. The mini-batch size of 64 is used. The initial learning rate is set to 0.1 and reduced to 0.01 and 0.001 at 50$\%$ and 75$\%$ of the total number of epochs, respectively. The training is still optimized by SGD algorithm with a momentum of 0.9 and weight decay of 0.0001.

Note that for the two CIFAR datasets, since data augmentation is turned off in the first round of training, to avoid overfitting, a dropout layer with a rate of 0.2 follows each convolutional layer except the first one. In the second round of training, we do not add dropout operations since data augmentation is backed on. For the SVHN dataset, given no data augmentation throughout the training process, dropout layers are added for both rounds of training.

\vspace{6pt}
\noindent\textbf{All-CNN.}
We also evaluate our approach with a typical sequential network (All-CNN) \cite{springenberg2014striving} on the two CIFAR datasets. In this model, max-pooling layer is replaced by regular convolutional layer with a stride of 2. We take the most advanced version of All-CNN, named as All-CNN-C in the original paper. Each block of this network contains two convolutional layers with a stride of 1 and one convolutional layer with a stride of 2.

We train the network using SGD optimizer with a momentum of 0.9 and weight decay of 0.001. The model is trained for 350 epochs, and the initial learning rate is set to 0.1. We adjust it by multiplying a fixed factor of 0.1 after 200, 250, and 300 epochs. To give reasonable comparison, in the second round of training, we only augment the data by horizontally flipping and translation of 5 pixels in maximum. The pre-processing steps include whitening and normalization.

\renewcommand{\arraystretch}{1.06}
\setlength{\tabcolsep}{1.0em}
\begin{table}[]
\resizebox{0.5\textwidth}{!}{%
\begin{tabular}{l|c|c|c}
\hline
\multicolumn{1}{c|}{Method} & CIFAR-10 & CIFAR-100 & SVHN \\ \hline
All-CNN \cite{springenberg2014striving} & 7.25 & 33.71 & - \\
All-CNN++  & \textbf{5.09} & \textbf{30.67} & -  \\ \hline
ResNet-110 (reported by \cite{huang2016deep}) & 6.41 & 27.22 & - \\
ResNet-152 (reported by \cite{huang2016deep}) & - & - & 2.01 \\
ResNet-110++  & \textbf{4.53} & \textbf{24.98} & - \\
ResNet-152++  & - & - & \textbf{1.64} \\
\hline
DenseNet-40 \cite{huang2017densely}  & 5.24 & 24.42 & 1.79  \\
DenseNet++  & \textbf{3.62} & \textbf{22.51} & \textbf{1.47} \\ \hline
\end{tabular}%
}
\vspace{0.5ex}
\caption{Image classification testing error rates (\%) for the well-known networks on the three datasets. \textquoteleft ++\textquoteright ~means the model is trained with our \emph{two-round training} approach. The better results are highlighted in bold.}
\label{classification}
\vspace{1ex}
\end{table}

\vspace{6pt}
\noindent\textbf{Analysis.} Table 1 lists the performance of the three networks trained with and without our approach. As can be seen, our \emph{two-round training} with \emph{data dropout} decreases the test error rates of the three networks on all the datasets, while the improvement on the two CIFAR datasets are greater than that on the SVHN dataset. This is because the images in the two CIFAR datasets contain more complicated scenarios, therefore, dropping \emph{unfavorable} training samples has a larger probability of removing disturbing features. 

The largest margin of improvement occurs on the All-CNN model. The reason can be attributed to the architecture, which is a sequential model in nature. Only the adjacent layers are connected, and there is no skip connection to feed different levels of features into subsequent layers. Therefore, this network is subject to the influence of \emph{unfavorable} samples. Nevertheless, ResNet and DenseNet can learn more distinguished features, hence they are more robust to \emph{unfavorable} samples. Similar interpretation also applies to the comparison between ResNet and DenseNet. Compared to DenseNet, our approach has achieved more performance advancement on ResNet. This holds true across all the three datasets. It indicates that \emph{data dropout} indirectly removes more disturbing features for ResNet, while relatively less for DenseNet due to its stronger ability of learning more distinguished features to suppress disturbing features. In fact, our approach indirectly proves that DenseNet outperforms ResNet, which has better performance than All-CNN. 

In addition, we illustrate the number of \emph{unfavorable} training samples in Table \ref{numdrop}. As can be seen, for the same training set, the amount is different for the three networks. \emph{Data dropout} locates more \emph{unfavorable} samples for All-CNN, while less for ResNet and DenseNet. This implies that our approach can improve inferior models by a larger margin. As visual examples, we in Figure 2 list several \emph{unfavorable} training samples that are picked from the CIFAR-10 dataset by the proposed data dropout scheme.    

\renewcommand{\arraystretch}{1.06}
\setlength{\tabcolsep}{1.0em}
\begin{table}[]
\center
\resizebox{0.5\textwidth}{!}{%
\begin{tabular}{l|c|c|c}
\hline
\multicolumn{1}{c|}{Method} & CIFAR-10 & CIFAR-100 & SVHN \\ \hline
All-CNN \cite{springenberg2014striving} & 1365 & 1419 & -  \\ \hline
ResNet-110 \cite{he2016deep} & 1220 & 1283 & - \\ \hline
ResNet-152 \cite{he2016deep} & - & - & 24261 \\ \hline
DenseNet-40 \cite{huang2017densely} & 1076 & 1149 & 19433 \\ \hline
\end{tabular}%
}
\vspace{0.5ex}
\caption{The number of \emph{unfavorable} training samples found for different models on the selected datasets.}
\label{numdrop}
\vspace{-1ex}
\end{table}

\begin{figure}
    \centering
    \includegraphics[width=0.4\textwidth]
    {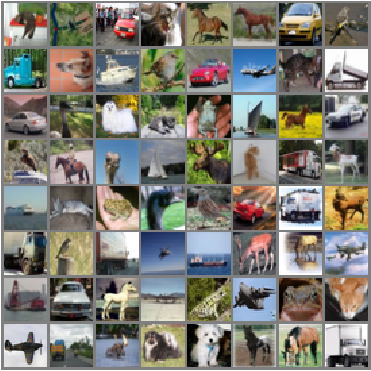}
    \caption{Example \emph{unfavorable} training images in the CIFAR-10 dataset.}
    \label{unfavorable}
\end{figure}

\subsection{Large Scale Image Classification}
To validate the effectiveness of the proposed approach for very large dataset, we conduct experiments on ImageNet \cite{deng2009imagenet} which is a benchmark dataset in image classification. We follow common practices \cite{krizhevsky2012imagenet, he2016deep, huang2017densely} to pre-process the images. Any image or its horizontal flip is randomly cropped to size $224\times224$. The per-pixel mean value is subtracted from each image and the standard color augmentation \cite{krizhevsky2012imagenet} is applied. We choose the ResNet-18 and ResNet-34 as the base networks, and train them by using the \textbf{Algorithm} as described in Section \ref{tr-train}. 

We use SGD optimizer to train both networks for 60 epochs. The momentum and weight decay are set to 0.9 and 0.001, respectively. We choose 0.1 as the initial learning rate and reduce it by multiplying 0.1 once the error has not decreased in the past three epochs. Following common practices, we compare validation errors (with 10-crop) only in Table \ref{imagenet_error}. As it shows, the data dropout and two-round scheme effectively boost the existing networks on the very large dataset. Similarly, in Table \ref{imagenet_numdrop}, we give the number of \emph{unfavorable} training samples that are removed from the original training set in ImageNet.

\begin{table}[t]
\begin{center}
\begin{tabular}{l|c|c}
\hline
& without \textbf{Algorithm} & with \textbf{Algorithm} \\
\hline
ResNet-18 \cite{he2016deep}& 27.88 & \textbf{24.26} \\
ResNet-34 \cite{he2016deep}& 25.03  & \textbf{21.97}  \\
\hline
\end{tabular}
\end{center}
\caption{Top-1 validation error (\%) of the two networks on the ImageNet dataset. The results in the second column are obtained by using conventional training method, and the ones in the last column are obtained based on our training scheme. The better results are highlighted in bold.}
\label{imagenet_error}
\end{table}

\renewcommand{\arraystretch}{1.06}
\setlength{\tabcolsep}{1.0em}
\begin{table}[]
\center
\begin{tabular}{l|c}
\hline
\multicolumn{1}{c|}{Method} & ImageNet  \\ \hline
ResNet-18 \cite{he2016deep} & 14655  \\ \hline
ResNet-34 \cite{he2016deep} & 13142 \\ \hline
\end{tabular}
\vspace{0.5ex}
\caption{The number of \emph{unfavorable} training samples removed from the original training set in ImageNet for the second round of training.}
\label{imagenet_numdrop}
\vspace{-1ex}
\end{table}

\subsection{Image Denoising}
As discussed in Section \ref{imgdenoising}, we re-evaluate DnCNN \cite{zhang2017beyond} without changing its default configurations. A noisy input is fed into the network and the output is the learned noise. We can obtain a clean image by subtracting the learned noise from the noisy input. Since our purpose is to validate the proposed \emph{data dropout} scheme and \emph{two-round training} approach, we only measure the effects for gray-scale image denoising with a known noise level. But it can be easily extended to color image denoising with random noise levels, since our approach is general and independent of models and applications.

\begin{figure*}[!htbp]
\begin{center}
\subfigure
{\includegraphics[width=0.077\textwidth]{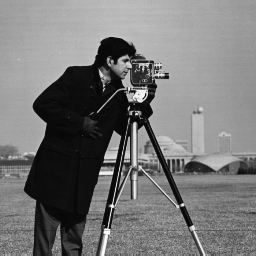}}
\subfigure
{\includegraphics[width=0.077\textwidth]{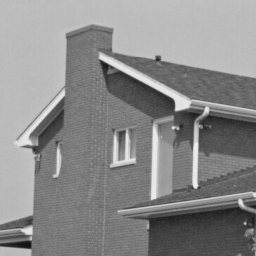}}
\subfigure
{\includegraphics[width=0.077\textwidth]{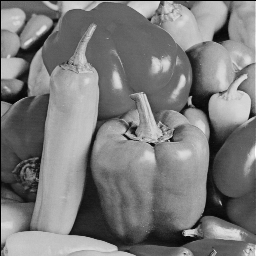}}
\subfigure
{\includegraphics[width=0.077\textwidth]{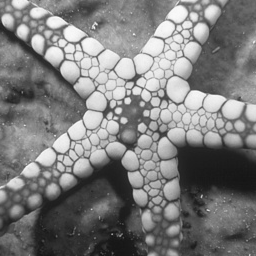}}
\subfigure
{\includegraphics[width=0.077\textwidth]{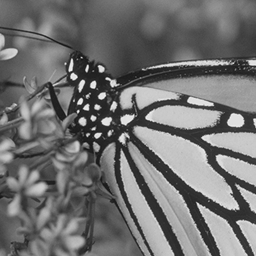}}
\subfigure
{\includegraphics[width=0.077\textwidth]{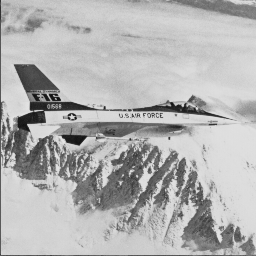}}
\subfigure
{\includegraphics[width=0.077\textwidth]{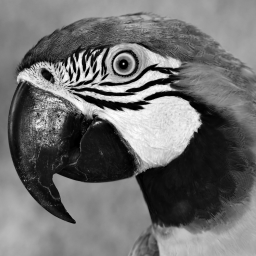}}
\subfigure
{\includegraphics[width=0.077\textwidth]{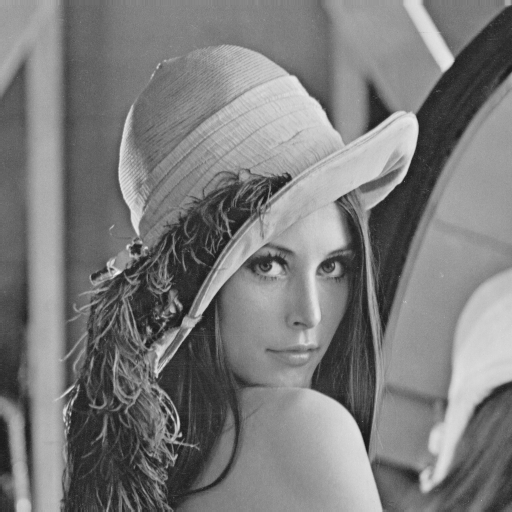}}
\subfigure
{\includegraphics[width=0.077\textwidth]{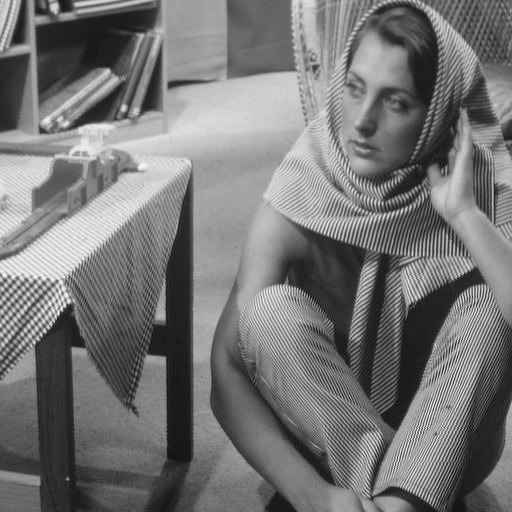}}
\subfigure
{\includegraphics[width=0.077\textwidth]{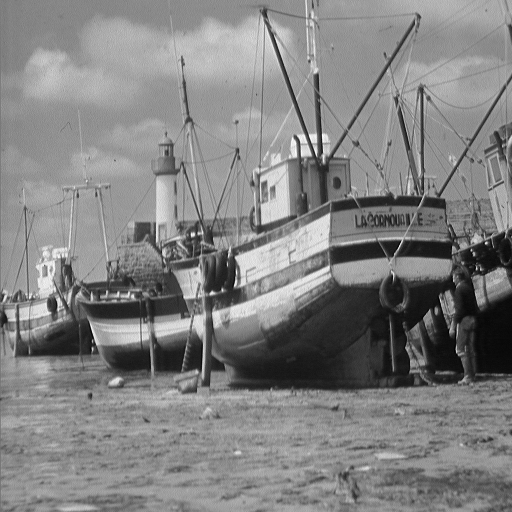}}
\subfigure
{\includegraphics[width=0.077\textwidth]{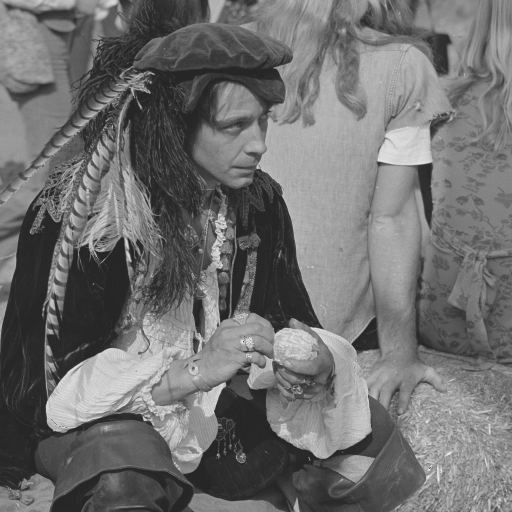}}
\subfigure
{\includegraphics[width=0.077\textwidth]{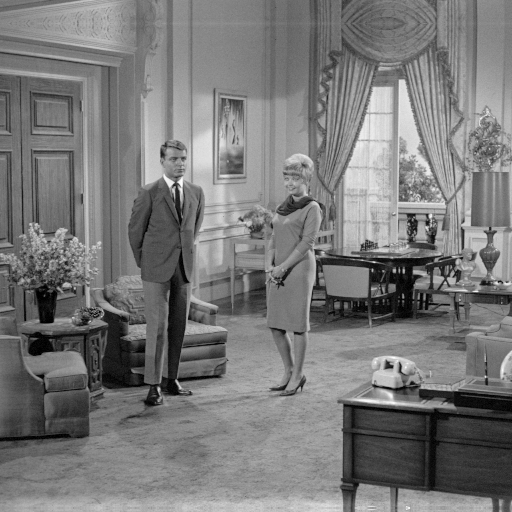}}
\caption{12 commonly used gray-scale images in image processing research.}
\label{12images}
\end{center}
\end{figure*}

\begin{table*}[htbp]
\center
\resizebox{0.95\textwidth}{!}{%
\begin{tabular}{|l|c|c|c|c|c|c|c|c|c|c|c|c|c|}
  \hline
  Images & C.man & House & Peppers & Starfish & Monar & Plane & Parrot & Lena & Barbara & Boat & Man & Couple & Average \\ \hline \hline
Noise Level & \multicolumn{13}{c|}{$\sigma=25$}   \\ \hline
DnCNN \cite{zhang2017beyond}& 30.18 & \textbf{33.06} & 30.87 & 29.41 & 30.28 & 29.13 & 29.43 & 32.44 & 30.00 & 30.21 & 30.10 & \textbf{30.12} & 30.436  \\\hline
DnCNN++ & \textbf{30.23} & 33.05 & \textbf{30.91} & \textbf{29.49} & \textbf{30.33} & \textbf{29.17} & \textbf{29.48} & \textbf{32.47} & \textbf{30.05} & \textbf{30.25} & \textbf{30.16} & 30.11 & \textbf{30.475}  \\\hline
\end{tabular}%
}
\caption{\emph{PSNR} comparison of the original DnCNN and our \emph{two-round trained} version DnCNN++ on the selected 12 gray-scale images, which are contaminated by a noise ($\sigma=25$) during testing. DnCNN and DnCNN++ are also trained based on the same noise level. Better \emph{PSNR} results are highlighted in bold.}
\label{denoising}
\end{table*}

\begin{figure*}[!htbp]
\begin{center}
\subfigure[\scriptsize Noisy / 20.38dB]
{\includegraphics[width=0.244\textwidth]
{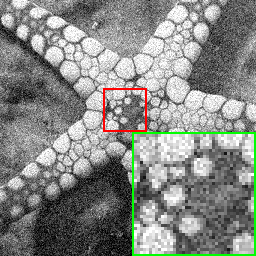}}
\subfigure[\scriptsize BM3D / 28.56dB]
{\includegraphics[width=0.244\textwidth]
{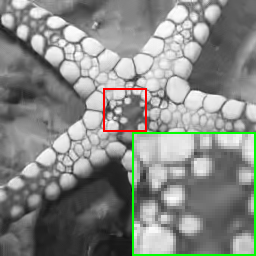}}
\subfigure[\scriptsize DnCNN / 29.41dB]
{\includegraphics[width=0.244\textwidth]
{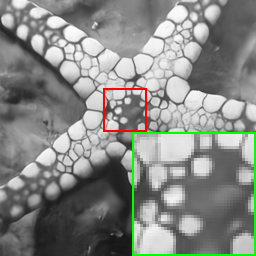}}
\subfigure[\scriptsize DnCNN++ / 29.49dB]
{\includegraphics[width=0.244\textwidth]
{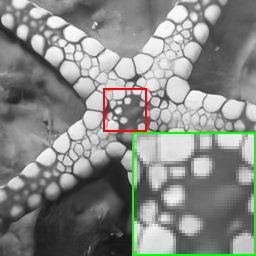}}

\caption{Example comparison of visual effects of different denoising results. The noisy input contains a specific noise $\sigma=25$.}
\label{visualdenosing}
\end{center}
\end{figure*}

We build the training set in a similar way as in \cite{zhang2017beyond} for the first round of training. 400 clean images are selected from Berkeley segmentation dataset (BSD500) \cite{MartinFTM01}. Each image is randomly cropped to a new image of size $180\times180$. Then $128\times1600$ patches of size $40\times40$ are further cropped from these images. Prior to training, additive Gaussian white noise with known kernel, namely $\sigma=25$, is added to the clean images to form noisy inputs. 12 commonly used gray-scale images in image processing research, as shown in Figure \ref{12images}, are used for testing purpose.

Note that in \cite{zhang2017beyond}, no validation set was used. However, in our approach, we need validation data to find \emph{unfavorable} training samples after the first round of training. Therefore, we use BSD68 dataset \cite{roth2009fields} for validation purpose and no cropping is taken. There is no common image between the training dataset and the BSD68 dataset. It is also important to highlight that, when evaluating $I_{loss}(x, x_{j})$, $x$ refers to a training image patch of size $40\times40$, while $x_{j}$ refers to a full size validation image.

Similarly, we adopt the \emph{MSRA} method \cite{he2015delving} to initialize network parameters for both rounds of training. We train the network using the Adam \cite{kingma2014adam} optimizer for 50 epochs. The initial learning rate is set to 0.001 for the first 30 epochs and adjusted to 0.0001 afterwards. Other default hyper-parameters of the Adam solver remain unchanged. The mini-batch size is set to 128. In the first round of training, no data augmentation is applied, whereas it is used in the second round of training.

We evaluate the trained model on the testing data, and compare the performance with the original DnCNN in Table \ref{denoising}. Here, the quality of restored images is measured in peak-signal-noise-ratio (PSNR), and larger values indicate better denoising results. It can be seen that adopting our training approach can increase the average PSNR by around 0.04dB for the given noise level, which is acceptable in image denoising. For the image \emph{\textbf{House}} and the image \emph{\textbf{Couple}}, our results are inferior to that of the original DnCNN. This is because when performing \emph{data dropout}, the influence of each training sample is estimated over the whole validation set, thus it cannot guarantee better performance for each individual testing sample. We illustrate the visual effects in Figure \ref{visualdenosing}. Besides the original DnCNN, we also list the visual effect of BM3D \cite{dabov2007image}, which is an image denoising method widely used in engineering.
As can be seen, the original DnCNN outperforms BM3D by a large margin, and our approach further improves DnCNN.

\section{Discussion}
\label{discuss}
In this section, we would like to provide several useful insights and discussions to help readers better understand our approach. 

One may argue that further training a network may increase the generalization accuracy, however, it could hardly bring a remarkable performance improvement if following the practices of the original papers of ResNet, DenseNet, and All-CNN. Those published results were actually measured based upon five times of independent running, and the best ones were reported. When our approach was not considered, we also attempted to improve those published results by increasing the training iterations, however, better testing results could not be obtained once the training had converged. On the other hand, although there are two rounds of training in our scheme, the second round of training is essentially different from further training or fine-tuning, because we train a model from scratch in the second round. In fact, the first round of training can be taken as a pre-processing step, which aims to optimize a given training dataset by reducing its size. 

Different networks do share several \emph{unfavorable} samples on the same training dataset, but the number is different because a more powerful network is robust to \emph{unfavorable} samples. That means a powerful network such as DenseNet will take fewer samples as \emph{unfavorable}, whereas an inferior model such as All-CNN-C will take more samples as \emph{unfavorable}. It is similar to human perception: a capable person usually takes surroundings positively while a pessimist may perceive more negative things from surroundings. 

Improving the computational efficiency for locating unfavorable samples could be an useful future work. For a specific model, assume the regular training time is $T$ (first round), our approach will totally cost $T$+$T_{s}$+$T_{d}$, where $T_{s}$ denotes the time of the second round of training and $T_{d}$ the time of data optimization. Here,  $T_{s}$ is less than $T$ because the second round of training is based on a reduced training set. $T_{d}$ is much less than $T$ because there is no back-propagation in data optimization. For a very large dataset, $T$ could be large, however, our work provides a practical way to take trade off. It can be used when domain-specific accuracy is highly desired.

\section{Conclusion}
\label{conclusion}

In this paper, to further boost performance of existing CNNs, we propose \emph{data dropout} scheme to optimize training data by removing \emph{unfavorable} samples. We theoretically analyze the criteria of \emph{data dropout} and point out it is convenient to apply in practice. To make use of the proposed scheme, we design a \emph{two-round training} approach which is general and can be easily integrated with existing networks and model configurations. Our experiments demonstrate the effectiveness of our approach for several well-known CNN models dealing with typical computer vision tasks.

\section*{Acknowledgments}

We gratefully acknowledge the support of NVIDIA Corporation with the donation of the Titan Xp GPU used for this research.

  \bibliographystyle{unsrt}
  \bibliography{Reference} 

\end{document}